\documentclass[conference]{IEEEtran}
\IEEEoverridecommandlockouts

\usepackage{cite}
\usepackage{amsmath,amssymb,amsfonts}
\usepackage{graphicx}
\usepackage{textcomp}
\usepackage{xcolor}
\usepackage{booktabs}
\usepackage{url}
\usepackage[hidelinks]{hyperref}
\usepackage{array}

\def\BibTeX{{\rm B\kern-.05em{\sc i\kern-.025em b}\kern-.08em
    T\kern-.1667em\lower.7ex\hbox{E}\kern-.125emX}}

\begin{document}

\title{HANSARD: A Reference Architecture for Forensic\\ Readiness, Runtime Witnessing, and Graded\\ Attribution in Autonomous Multi-Agent AI Systems}

\author{

  \IEEEauthorblockN{
    Christos~Sardianos\IEEEauthorrefmark{1},
    Iliana~Pla\IEEEauthorrefmark{1},
    Vasilis~Efthymiou\IEEEauthorrefmark{1},
    Iraklis~Varlamis\IEEEauthorrefmark{1},
    Thomas~Lagkas\IEEEauthorrefmark{2},\\
    Panagiotis Sarigiannidis\IEEEauthorrefmark{3},
    Georgios~Th.~Papadopoulos\IEEEauthorrefmark{1}\IEEEauthorrefmark{4}\\
    \thanks{This work has received funding from the European Union's Horizon Europe research and innovation programme under Grant Agreement GANNDALF (No. 101167951) and SALUS (No. 101225719).}% <-this % stops a space
    }

    \IEEEauthorblockA{
        \IEEEauthorrefmark{1}Department of Informatics and Telematics, Harokopio University of Athens, Athens, Greece\\
        \IEEEauthorrefmark{2}Department of Informatics, Faculty of Science, Democritus University of Thrace, Kavala, Greece\\
        \IEEEauthorrefmark{3}Department of Electrical and Computer Engineering, University of Western Macedonia, Kozani, Greece\\
        \IEEEauthorrefmark{4}Archimedes, Athena Research Center, Athens, Greece\\
    }

    \IEEEauthorblockA{
        Emails: \{sardianos,ilianapla,vefthym,varlamis,g.th.papadopoulos\}@hua.gr, tlagkas@cs.duth.gr, psarigiannidis@uowm.gr
    }
    
}

\maketitle

\begin{abstract}
Autonomous multi-agent systems nowadays act in finance, software supply chains, and security operations. Already, the first largely AI-orchestrated intrusion campaigns have been reported. Yet, when such a system causes harm, no method can robustly establish what happened, what caused it, or who is accountable. This is because provenance forensics works at the wrong abstraction, formal causality assumes the causal model, and agent auditing trusts self-recording. The target failure mode is, thus, attribution laundering, i.e., spreading an act across redundant agents until none is a but-for cause. Worse, the record is produced by the suspects, which comprises the assumption adopted throughout this work. Agents may therefore anticipate the investigation and the part of logging infrastructure may itself collude. In this paper, HANSARD is proposed, a reference architecture treating accountability as a life-cycle property. First, a readiness profile sealed before operation bounds what later findings may claim. Second, capturing at five choke points beyond the agents' reach makes omissions detectable, not only tampering. Third, a typed PROV-DM-aligned causal graph accrues as the system runs, and three indicators read it live to gate oversight without adjudicating. Fourth, post-incident replay yields contingent effects under the modified Halpern-Pearl definition, together with a compensation-set size. Finally, a synergy residual measures harm due to the combination rather than to individuals, making laundering visible. Cause, responsibility and accountability are then reported separately, each capped by an evidentiary tier, while a future research agenda is also provided.
\end{abstract}


\section{Introduction}

Language models with tools, memory and delegation authority are increasingly deployed as interacting agent populations acting for a principal, while a growing literature documents the cost. In particular, agents remain measurably influenced by content labeled untrusted \cite{provsens}, coordinate through channels leaving no observable trace \cite{collusion}, and emit rationales that do not track the determinants of their behavior \cite{turpin}, degrading further under optimization pressure against a monitor \cite{cotmonitor}. Consequently, work on agent auditability has formed in response \cite{auditable,capchain,agentbom,causalreplay}, prompted by deployment outpacing accountability and the absence of an identity standard tracing authorization to a human principal \cite{aiidentity}. Ultimately, both lines converge, where the evidence explaining a harmful outcome is generated inside the system under question, while provenance-based forensics \cite{backtracker,beep} operates where that influence is invisible.

\emph{Agentic crime} denotes unlawful or actionable harm brought about through autonomous agents rather than a human hand, criminal intent aside. Its setting is one in which autonomy is distributed by construction, as in open multi-agent architectures that share learning and control across many participants \cite{papadopoulos2021}. Where such harm arises within an agent population delegating and writing to shared memory across organizational boundaries, three main questions current practice conflates must be resolved: \emph{what occurred} (reconstruction), \emph{what brought it about} (causal attribution), and \emph{who is answerable} (accountability). Explainability addresses the first two only in part, since post-hoc multi-modal accounts explain a model's output rather than an event \cite{rodis2024}, while counterfactual explanation, closest among them to a causal claim, is defined over an input rather than over a distributed execution \cite{evangelatos2026}. Moreover, formal responsibility attribution addresses the third \cite{munajib}, without tracking the judgments a tribunal would make \cite{saxena}. Overall, each question carries a distinct evidentiary standard, and only the third is what regulators and courts require.

The deficiency described above is not merely technical, since regulation of high-risk AI already presupposes an evidentiary capability that has not been built, requiring operational records to be retained and serious incidents reported within short deadlines, with the causal link characterized where one is established or reasonably likely \cite{aiact}. Such duties assume an operator controlling its own logs, not a composition of agents from several organizations, where no party holds the whole record and the causal chain crosses trust boundaries. On the liability side, the instrument that would have eased a claimant's burden through a rebuttable presumption of causation was withdrawn \cite{aild}. Two obstacles compound this. Evidence from proprietary systems is routinely withheld from the parties it is used against \cite{wexler} and forensic readiness, established for conventional IT \cite{rowlingson}, has not reached agentic systems, where the monitored can model the monitor. The challenge is therefore not further obligation, but a technical account of what such a system must record and what a reconstruction may legitimately claim.

The current work focuses on and advances a central position that post-hoc accountability for multi-agent AI fails at a juncture no current design accounts for, namely that the record is produced by the suspects. Specifically, current designs assume an honest recorder, whereas an adversary anticipating reconstruction acts on the record itself; accountability must therefore be designed against that adversary and cannot initiate with incident response. To this end, this paper introduces the following contributions towards this direction: a) The first characterizes the multi-agent forensic problem and identifies \emph{attribution laundering} (i.e., dispersing a harmful act over interchangeable agents,  until no individual counterfactual effect remains) as its signature failure mode. b) The second is \emph{HANSARD}, a life-cycle architecture comprising a sealed readiness profile, an out-of-band witness boundary, a typed causal graph accrued at runtime, and a post-incident counterfactual pipeline. c) The third applies the gap function of cooperative game theory to counterfactual harm effects, yielding an instrument for laundering and a structural proxy computable without replay. d) The fourth is a proportionality discipline between claims and evidence, as evidentiary tiers whose attainability is fixed before deployment.

The remainder of the paper is organized as follows. Section~\ref{sec:related} reviews provenance forensics, actual causality, and agent auditability. Section~\ref{sec:threat} discusses the threat model and requirements. Section~\ref{sec:arch} outlines the architecture. Section~\ref{sec:tiers} provides the evidentiary tiers and readiness levels. Section~\ref{sec:agenda} summarizes the limitations and a future research agenda. Finally, Section~\ref{sec:conclusion} concludes the manuscript.

\section{Background and Related Work}
\label{sec:related}

\textbf{Provenance-based forensics.} Causal reconstruction from audit data began with BackTracker \cite{backtracker} and execution partitioning for \emph{dependency explosion} \cite{beep}, while it is evadable by substituting rare system actions with common ones \cite{provninja}. It materializes at the operating-system layer, whereas agentic influence propagates through natural-language content with no kernel-observable edge; a mismatch of abstraction, not of coverage.

\textbf{Actual causality and responsibility.} In the structural-model account \cite{hp2015}, $X$ is a cause if, holding a \emph{witness set} $W$ fixed, varying $X$ alters the outcome; the degree of responsibility $1/(1+k)$ grades this by the minimal number $k$ of other variables that must be altered to render $X$ critical \cite{chockler}, and extends to probabilistic multi-agent games \cite{munajib}. Deciding causality is $\Sigma_2^P$-complete under the original definition \cite{complexity}, $D_1^P$-complete (NP-complete for one variable) under the modified one \cite{hp2015}. Two limitations are material, namely the causal model is \emph{presupposed} (never recovered from contaminated evidence), and a human-subject study found no formal measure aligning consistently with human judgments \cite{saxena}.

\textbf{Agent auditability and attribution.} Recent literature addresses agent observability, including \emph{Auditable Agents} \cite{auditable}, CapChain \cite{capchain}, Agent-BOM \cite{agentbom} and Causal Agent Replay \cite{causalreplay}; no production identity standard though traces authorization to an originating human principal \cite{aiidentity}. The present work
departs from that literature in two respects: a) That work is non-adversarial, trusting capture, whereas the setting here is a system built, or induced, to defeat investigation, and b) It is retrospective, whereas forensic readiness provisions evidence beforehand \cite{rowlingson} and runtime verification checks executions as they run \cite{leucker}, neither having reached agentic systems, where the monitored can model the monitor.

\textbf{Evidence and admissibility.} Error rate is one non-exclusive \emph{Daubert} factor \cite{daubert}, considered in this work as a design target rather than a legal test, the EU regimes at issue leaving admissibility to national procedural law. Defendants routinely cannot inspect the proprietary systems generating evidence against them \cite{wexler} and tamper-evident logging \cite{rfc9162} proves integrity, never completeness, a limit inherited from secure logging on untrusted machines \cite{schneier}. Moreover, a self-reported rationale is not evidence of a cause \cite{turpin}, monitorability being fragile under optimization pressure against the monitor \cite{cotmonitor}; self-reports are therefore artifacts requiring explanation, not testimony.

\section{Threat Model and Requirements}
\label{sec:threat}

Agents from possibly distinct providers exchange messages, invoke tools, and share memory on behalf of a human principal, who supplied only a high-level objective. Harm $\varphi$ has occurred and an external investigator must produce findings that withstand scrutiny. Three adversary classes are distinguished, as follows: A1) The external attacker, who compromises agents through untrusted content (indirect prompt injection, tool-metadata poisoning, memory poisoning) without touching the infrastructure, as demonstrated in deployed applications \cite{asimopoulos2026}. A2) The anticipating agent, who acts in the knowledge that it is monitored currently and will be reconstructed subsequently, by emitting exculpatory rationales, routing influence through benign-appearing intermediaries, employing covert channels \cite{collusion}. Its most effective strategy, designated in this work `attribution laundering', distributes the constituent acts across redundant agents so that under a na\"ive but-for counterfactual no agent is critical. A3) The colluding recorder, who controls part of the logging infrastructure and can omit, reorder, or fabricate entries.

Table~\ref{tab:threat} maps the above capabilities onto the requirements they induce and the respective countermeasures. (A3) cannot be defeated within one trust domain, only converted into a detectable inconsistency \emph{across} domains, whence federation is an architectural requirement, not an optional mode.

\begin{table}[!t]
\caption{Adversary capabilities, the requirements they induce, and architectural countermeasures}
\label{tab:threat}
\centering
\scriptsize
\renewcommand{\arraystretch}{1.0}
\begin{tabular}{@{}m{0.25\columnwidth}>{\centering\arraybackslash}m{0.13\columnwidth}m{\dimexpr0.62\columnwidth-4\tabcolsep\relax}@{}}
\toprule
\textbf{Capability} & \textbf{Req.} & \textbf{Countermeasure (component)} \\
\midrule
Injects instructions via untrusted content (A1) & R1 & Provenance-labeled context segmentation, labels assigned at the witness boundary, not by the agent (Section~\ref{sec:acg}) \\ \midrule
Emits misleading rationale or uses a covert channel (A2) & R1, R3 & Self-reports excluded from the causal model; content-blind mandatory capture at W$_3$, so unexplained influence raises $\rho$ (Section~\ref{sec:acg}) \\ \midrule
Attribution laundering (A2) & R4, R5 & Contingent effects with minimal witness-set search; synergy residual $\rho$ and its runtime shadow (Section~\ref{sec:cf}, Section~\ref{sec:runtime}) \\ \midrule
Omits or reorders entries; controls a trust domain or waits out retention (A3) & R2, R6, R7 & Per-channel sequence numbers and hash chaining yield completeness proofs, not merely integrity proofs, and gaps raise live alarms (Section~\ref{sec:capture}, Section~\ref{sec:runtime}); cross-principal corroboration, with tier E3 required to name a principal (Section~\ref{sec:tiers}); commitment retention decoupled from content retention (Section~\ref{sec:fed}) \\
\bottomrule
\end{tabular}
\end{table}

Following the above discussion, a list of requirements is formed, where each the obligation that one element of the foregoing analysis imposes: R1)~Capture occurs outside the agents' sphere of influence. R2)~The record supports channel-relative completeness, not only integrity, proofs. R3)~Dependency structure is recovered from evidence rather than presupposed, and interventions evaluated by replay rather than against a hand-specified structural model. R4)~Attribution is graded and remains informative under over-determination. R5)~Every claim carries a stated evidentiary basis and error characterization. R6)~Reconstruction spans trust domains without disclosing proprietary internals. R7)~The same sealed record serves in-flight oversight and post-incident examination, where runtime consumers are read-only with respect to evidence and every intervention they trigger is itself witnessed.

\begin{figure*}[!t]
\centering
\includegraphics[width=1.00\textwidth]{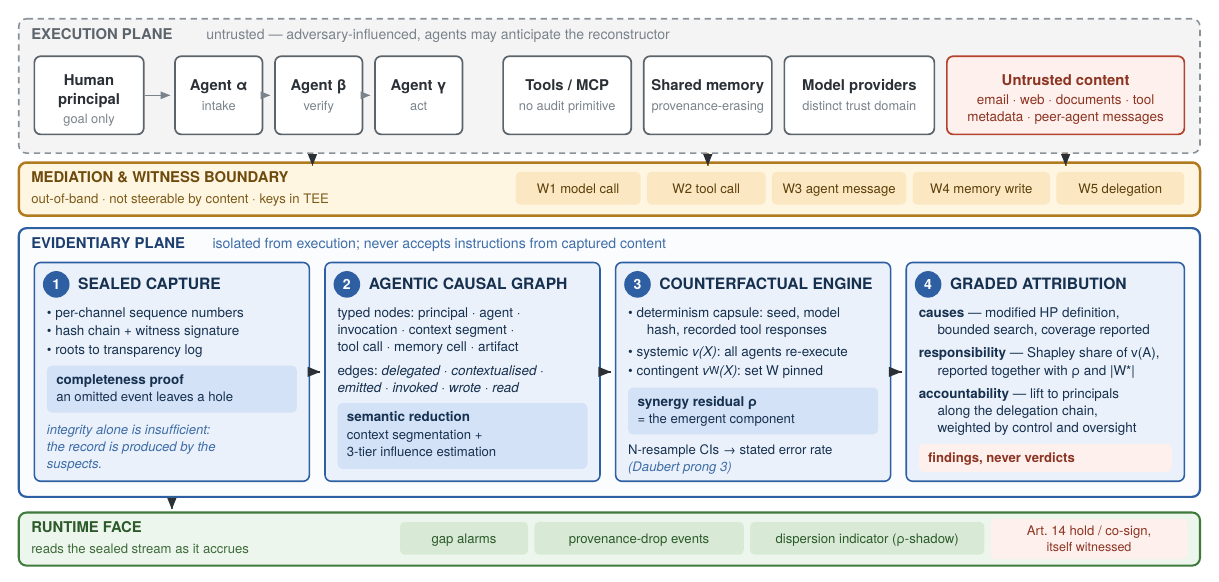}
\caption{The HANSARD reference architecture. The witness boundary is the only sanctioned path between the untrusted execution plane and the evidentiary plane; stages 1-4 constitute the post-incident examination and the runtime face reads the same sealed stream as it accrues. The readiness profile (Section~\ref{sec:ready}) and federation (Section~\ref{sec:fed}) are omitted from the figure.}
\label{fig:arch}
\end{figure*}

\section{The HANSARD Architecture}
\label{sec:arch}

The introduced HANSARD architecture is graphically illustrated in Fig.~\ref{fig:arch}, comprising a strict separation between an \emph{execution plane}, untrusted in its entirety, and an \emph{evidentiary plane}, which never accepts instructions from captured content; the two communicating solely through a mediation and witness boundary. Additionally, the evidentiary plane has three temporal faces of one record, namely the readiness profile, the runtime face, and the examination pipeline of stages~1-4.

\subsection{The readiness profile}
\label{sec:ready}
Before the system acts at all, the deployer seals a \emph{readiness profile}, including the enumeration of mediated and known unmediated channels; the determinism-capsule provisions of Section~\ref{sec:cf}; the indicator thresholds and escalation hooks of Section~\ref{sec:runtime}; and the retention and federation commitments of Section~\ref{sec:fed}. The profile is itself witnessed and published, so a later investigation begins from a signed statement of what the deployment promised to be able to answer, and a tribunal can distinguish evidence never provisioned from evidence gone missing.

\subsection{Sealed capture}
\label{sec:capture}
Observation occurs at five choke points an agent cannot bypass without failing to act at all, namely model invocation (W$_1$), tool call (W$_2$), inter-agent message (W$_3$), memory write (W$_4$) and delegation (W$_5$). Witnesses are not agents, hold signing keys in a trusted execution environment, and are not steerable by the content they record. Each maintains a monotonic per-channel sequence number and hash-chains its entries; roots are published continuously to an append-only transparency log \cite{rfc9162}.

Integrity proofs establish that what is present was not altered; under capability (A3) the pertinent question is what is \emph{absent}. Per-channel sequence numbering converts an omission on a mediated channel into a gap that consistency proofs expose, rendering absence evidentially informative. The guarantee is channel-relative (unmediated channels remain a stated residual assumption (Section~\ref{sec:agenda})), but it is the concrete sense in which securing storage, while trusting capture, is the wrong threat~model.

\subsection{The agentic causal graph}
\label{sec:acg}
The evidentiary plane compiles observations into an \emph{agentic causal graph} (ACG), which comprises a typed attributed digraph whose nodes are principals, agent instances, model invocations, context segments, tool calls, memory cells, messages and artifacts, and whose edges are typed by influence mode (namely, `delegated', `contextualized', `emitted', `invoked', `wrote' and `read'). Additionally, the vocabulary aligns with the responsibility relations of PROV-DM \cite{provdm}, so `actedOnBehalfOf' chains are first-class and the graph serializes into an existing standard. The graph is built incrementally as witnessed entries arrive; the post-incident graph is the runtime graph, frozen when an examination opens.

Two architectural decisions need to be highlighted. First, provenance labels on context segments are assigned by the witness boundary at ingestion and immutable thereafter, since in-context labeling is not self-enforcing, i.e., agents remain measurably influenced by content marked untrusted \cite{provsens}. Second, an agent's account of its own reasoning enters the graph as an artifact with no causal edges, where admitting self-reports as causal edges would import the very unreliability \cite{turpin} the architecture exists to control.

\textbf{Semantic dependency reduction.} The unpruned ACG is intractable, since every token in a shared context is potentially a cause of every subsequent action; a dependency explosion more severe than the one that motivated execution partitioning \cite{beep}. Context is therefore segmented by provenance (each span injected from a distinct source at a given time is one unit, delimited by the mediation layer) and candidate edges pruned by a three-tier estimator: lexical and embedding overlap, then attention or attribution scoring where white-box access exists, are triage deciding where counterfactual budget is spent; only the third tier, counterfactual ablation, yields an evidentially usable claim.

\subsection{The runtime face}
\label{sec:runtime}
The quantities of Section~\ref{sec:cf} require replay; three indicator families need only the accruing record. a) \emph{Gap alarms}: A per-channel sequence discontinuity (Section~\ref{sec:capture}) is detectable the moment the succeeding entry arrives, converting an (A3) omission into a live signal. b) \emph{Provenance-degradation events}: A memory write depending on an untrusted-labeled segment while citing no witnessed verification act is the moment content crosses from untrusted to internally trusted state, and W$_4$ observes it as it happens. c) \emph{Structural dispersion}: A shadow of the synergy residual; when the causal frontier of a pending consequential action spans interchangeable instances such that no single upstream node is a cut vertex of the streaming ACG, every singleton but-for effect is predictably near zero, so laundering-shaped structure is flagged from topology alone, before the harm completes.

Indicators do not adjudicate; they gate. Crossing a threshold declared in the readiness profile triggers oversight hooks at the mediation boundary; holding an action pending human co-signature, the operative mechanism of the Article~14 oversight duty \cite{aiact}, or quarantining a memory cell. The runtime face is read-only with respect to evidence (R7) and every intervention it triggers is itself witnessed. The caveat is structural, where optimization pressure against a monitor degrades the behavior monitored \cite{cotmonitor}, so thresholds are fixed in the sealed profile, not adapted silently; the runtime face is a sensor, not a guard, and its calibration remains open (Section~\ref{sec:agenda}).

\subsection{Counterfactual engine}
\label{sec:cf}
Let $T$ denote the recorded trace, $\varphi$ the harm predicate, and $\mathcal{A}$ the agent set. A \emph{determinism capsule} records seed and sampling parameters, model version hashes and all tool responses, so replay runs against a recorded environment with side effects stubbed from the trace; where exact reproduction is unattainable, estimation proceeds over $N$ resampled replays with reported intervals. Write $P_T$ for the replay distribution seeded from the capsule and $do(X{\leftarrow}\bot)$ for suppression of an event or agent $X$ (the invocation is elided; downstream consumers observe absence, not an error); for $S \subseteq \mathcal{A}$, $v(S)$ suppresses every member jointly, extending $v$ to a set function on $\mathcal{A}$.

The \emph{systemic effect} of $X$ is $v(X) = P_T(\varphi) - P_T(\varphi \mid do(X{\leftarrow}\bot))$, every agent outside $X$ being free to re-execute and compensate. Since $P_T(\varphi) \approx 1$ under faithful replay, $v$ is effectively the probability that suppressing $X$ averts the harm; the but-for quantity proper, near zero for every singleton under redundancy, which is precisely the laundering problem. With $P_W$ for $P_T$ under a set $W$ of events pinned at their recorded values, the \emph{contingent effect} is $v_W(X) = P_W(\varphi) - P_W(\varphi \mid do(X{\leftarrow}\bot))$. The \emph{modified} HP definition \cite{hp2015} is adopted, under which a contingency may only be set to its actual value (precisely the pinning above) so that $X$ is an actual cause of $\varphi$ if $X$ occurred and some $W$ renders $v_W(X)$ \emph{significant}, meaning the lower bound of its $95\%$ interval over $N$ replays exceeds a threshold $\tau$ declared with every finding. The modified definition yields the favorable complexity noted above, so the binding cost is replay batches, not combinatorics. It also renders the Chockler-Halpern degree of responsibility $1/(1+k)$ inapplicable, since $k$ counts variables set to values \emph{other} than actual \cite{chockler} and in the current conceptualization there are none. The \emph{compensation-set size} $|W^{\ast}|$ (i.e., the number of agents whose non-action had to be pinned) is therefore reported as a distinct structural measure, not a degree of responsibility. Search runs to a declared bound $k_{\max}$ with coverage reported, so a null result reads ``no cause found within $|W| \le k_{\max}$'', never exoneration; reported causes carry the sampling error of the $N$-replay estimate and are not certificates.

\textbf{Synergy residual.} Normalizing the gap function of the effect game yields:
\begin{equation}
\rho \;=\; \frac{v(\mathcal{A}) - \sum_{a \in \mathcal{A}} v(\{a\})}{v(\mathcal{A})}, \qquad v(\mathcal{A}) > 0,
\label{eq:rho}
\end{equation}
where $\rho$ is a signed index in $(-\infty, 1]$, reported as not-applicable when $v(\mathcal{A}) \le 0$. It is a comparison, not a decomposition (harm removable by dismantling the entire agent set against harm removable one agent at a time) and $1-\rho$ must not be construed as harm attributable to individuals. Values near $1$ indicate that no member is individually decisive and the composition produced the harm; values at or below $0$ indicate the converse, as in a serial pipeline where each member could individually have averted~it. This renders laundering legible, i.e., an adversary distributing an act across redundant agents drives each $v(\{a\})$ toward zero, which \emph{raises} $\rho$ rather than dissolving responsibility. A high $\rho$ is a positive finding; what a regulator concerned with system-level obligations, or a claimant confronting a many-hands defense, requires on the record. The many-hands problem is long established; the residual makes its multi-agent form measurable. The statistic is the standard gap function of a cooperative game; the contribution lies in its forensic interpretation.

\subsection{Graded attribution}
The final stage produces three separated outputs: a) \emph{Causes} are events for which a witness set was found within $k_{\max}$, each reported with its compensation-set size $|W^{\ast}|$. b) \emph{Responsibility} apportions $v(\mathcal{A})$ by Shapley value \cite{munajib}, reported alongside $\rho$. The two are complementary, not competing; Shapley efficiency assigns all of $v(\mathcal{A})$ to members and $\rho\,v(\mathcal{A}) = \sum_i (\phi_i - v(\{i\}))$ is the portion each agent owes to interaction rather than standalone effect. c) \emph{Accountability} lifts from agents to principals along `actedOnBehalfOf' chains, weighted by foreseeability and by control; who could have intervened, the operative question under the Article~14 human-oversight duty \cite{aiact}. The separation is itself a claim, where the literature routinely conflates the three, though the evidentiary requirements differ at each step; since no formal responsibility measure tracks human judgment reliably \cite{saxena}, HANSARD emits the evidence and the measure, never a verdict.

\subsection{Cross-principal federation}
\label{sec:fed}
In a realistic incident the evidence is split across a model provider, an orchestration framework, a tool provider, and the deployer, with MCP offering no audit primitive and A2A only a correlation identifier \cite{aiidentity}. HANSARD requires each trust domain to publish Merkle roots of its witness log continuously and to disclose content only under legal process, with inclusion proofs. In this respect, a log can then be authenticated without publishing the model that produced it, partly answering the trade-secret objection \cite{wexler}, and \emph{commitment} retention, low-cost and durable for years, is decoupled from \emph{content} retention, whose Article~26(6) floor of six months is short relative to intrusion dwell times and litigation \cite{aiact}.

\section{Evidentiary Tiers and Readiness Levels}
\label{sec:tiers}

Every finding carries a tier (as explained in Table~\ref{tab:tiers}) under certain promotion rules, namely causation requires at least (E2), naming a principal at least (E3). (E1) evidence is low-cost and will predominate in any realistic investigation; letting it support attribution to a named party is how forensic architectures produce high-confidence misattributions.

\begin{table}[!t]
\caption{Evidentiary tiers and what each may license}
\label{tab:tiers}
\centering
\scriptsize
\renewcommand{\arraystretch}{1.0}
\begin{tabular}{@{}>{\centering\arraybackslash}m{0.09\columnwidth}m{0.36\columnwidth}m{\dimexpr0.55\columnwidth-4\tabcolsep\relax}@{}}
\toprule
\textbf{Tier} & \textbf{Basis} & \textbf{May license} \\
\midrule
E0 & Correlational narrative over the trace & Hypothesis generation only; not evidence \\ \midrule
E1 & Sealed trace, integrity \emph{and} completeness proofs, one witness & Description of what was recorded; remediation advice \\ \midrule
E2 & (E1) plus counterfactual support, with stated interval and $k_{\max}$ coverage & A claim of causation \\ \midrule
E3 & (E2) corroborated by independent witnesses in distinct trust domains & A claim naming a principal \\
\bottomrule
\end{tabular}
\end{table}

The tiers are gated by what was provisioned before the incident, graded as readiness levels: L0) No sealed capture, caps every finding at (E0). L1) Witnessed channels enumerated, capsules provisioned, thresholds declared; makes (E1) and (E2) attainable. L2) Adds standing cross-domain federation commitments (Section~\ref{sec:fed}), the precondition for (E3). No post-incident diligence can retrofit a tier the profile did not provision; the precise sense in which accountability is a life-cycle rather than an investigative property. A reconstruction attaining only (E1) has, thus, not failed silently; it has reported that the evidence cannot support a causal claim.

\section{Limitations and Future Research Agenda}
\label{sec:agenda}

What is proposed in this study is an architecture, not a system; several components remain open. \textbf{Counterfactual validity.} Determinism capsules are effective where the model is self-hosted or exposes seeds; for closed APIs subject to silent version drift the intervention distribution may not be estimable at all, and the sound response is to withhold the (E2) tier rather than report a meaningless interval. HANSARD also substitutes a replay oracle for the structural equations HP causality presupposes; sufficient to evaluate an intervention, not the model-theoretic conditions. Establishing when replay is valid is the largest single gap. \textbf{Validating $\rho$ and calibrating its shadow.} The residual is uncalibrated and its behavior on mixed structures (partly redundant, partly serial) is unstudied. The dispersion indicator inherits both problems and adds a false-positive burden, where redundancy engineered for availability is laundering-shaped by construction, so an uncalibrated threshold either saturates operators with escalations or never fires. Both need evaluation against known ground truth, mindful that formal measures diverge from human judgment \cite{saxena}. \textbf{Capture integrity.} A witness observes only what traverses it, where an effect achieved through an unwitnessed channel (an unmediated side effect, a covert encoding \cite{collusion}) leaves no discontinuity to expose, so enumerating a deployment's unwitnessed channels is a prerequisite to any completeness claim. \textbf{Cost.} Mediating five choke points, evaluating indicators in stream and executing $N$-replay ablations costs latency on the execution plane and compute on the evidentiary plane \cite{auditable}; whether sealed capture is affordable at production throughput, and how to allocate a fixed replay budget, is unaddressed. \textbf{Standardization.} None of this composes across organizations without an interchange format; the PROV-DM-aligned ACG serialization \cite{provdm} is the natural candidate, and the missing audit primitive in current agent protocols the natural starting point.

\section{Conclusion}
\label{sec:conclusion}

This paper has addressed the forensic vacuum that opens when harm is brought about by autonomous multi-agent systems whose record is produced by the parties under investigation, and whose audit, in current practice, begins only once the harm is done. Four contributions/outcomes have been presented. First, attribution laundering was identified as the signature failure mode of such systems, dispersing a harmful act over interchangeable agents, until no individual counterfactual effect remains. Second, HANSARD was specified as a life-cycle reference architecture joining a sealed readiness profile, an out-of-band witness boundary over five choke points, a typed causal graph accrued at runtime with three live indicators, and a post-incident counterfactual pipeline. Third, the gap function of cooperative game theory was adapted to counterfactual harm effects as a synergy residual that renders laundering visible, together with a structural proxy computable without replay. Fourth, evidentiary tiers gated by readiness levels were defined, bounding what any reconstruction may license and making evidence never provisioned a matter of record rather than a silent omission. A future research agenda identified replay validity, calibration of the residual, unwitnessed channels, and cost as the principal open problems.

\section*{Acknowledgment}

The authors disclose that the Claude AI assistant (Anthropic) supported the literature search underlying Section~\ref{sec:related} and generated the schematic of Fig.~\ref{fig:arch}. All such content was verified and edited by the authors, who accept full responsibility for this work.

\bibliographystyle{IEEEtran}
\bibliography{references}

\end{document}